\documentclass[10pt,sigconf,nonacm]{acmart}
\setcopyright{none}

\begin{document}

\title{\raisebox{-0.15\height}{\includegraphics[height=1.2em]{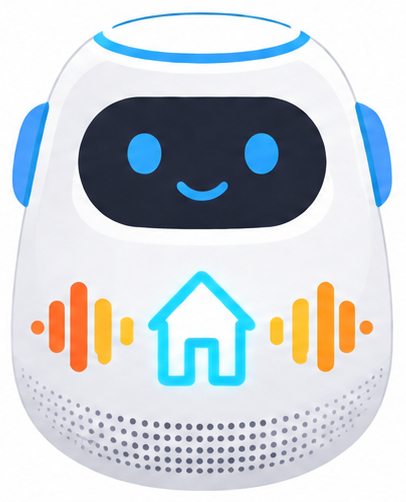}}\hspace{0.3em}Homebot: A Personal AI Agent for Conversational Home Assistance and Automation}

\author{Shengyuan Ye$^\blacklozenge$, Yixin Zhang$^\lozenge$, Han Liang$^\blacktriangle$, Liekang Zeng$^\triangle$, Jiangsu Du$^\blacktriangle$, Mu Yuan$^\triangle$}
\affiliation{
  \country{$^\blacklozenge$Power Dispatching and Controlling Center, Guangdong Power Grid Co., Ltd., Guangzhou, China}
}
\affiliation{
  \country{$^\lozenge$Independent Researcher}
}
\affiliation{
  \country{$^\blacktriangle$School of Computer Science and Engineering, Sun Yat-sen University, Guangzhou, China}
}
\affiliation{
  \country{$^\triangle$Department of Information Engineering, The Chinese University of Hong Kong, Hong Kong SAR, China}
}
\email{Contact: brandonye@foxmail.com}
\email{Website: https://ysyisyourbrother.github.io/homebot}

\renewcommand{\shortauthors}{Ye et al.}

\begin{abstract}
\texttt{Homebot} is a locally deployable AI agent for conversational household assistance and automation. It accepts voice and instant-messaging requests through a shared runtime that combines language-model responses with registered tools and task-specific skills. The design separates common request processing from session ownership: messaging history remains scoped to a channel and chat, whereas voice interaction is bounded by wake-word activation. For hands-free use, \texttt{Homebot} combines local wake-word detection, streaming speech recognition and synthesis, and an explicit dialogue-state protocol for ending, following up, or continuing a conversation. Clear channel, tool, and skill contracts support practical customization for household use. The source code is publicly available at \url{https://github.com/ysyisyourbrother/homebot}.
\end{abstract}

\maketitle

\section{Introduction}

Recent advances in language models~\cite{xu2026deepseek, team2026kimi} have expanded agent systems from text generation toward iterative reasoning and interaction with external environments and tools~\cite{yao2023react,schick2023toolformer}. This direction has produced an active ecosystem of open-source personal-agent software, including OpenClaw, nanobot, and Hermes Agent~\cite{openclaw,nanobot,hermesagent}. These systems make language-model capabilities more accessible for individual workflows, but household assistance poses a distinct interaction and customization problem.

Home activities often require help when a person is cooking, cleaning, or otherwise away from a keyboard \cite{king2024sasha, ye2026venus}. In these settings, voice interaction is embedded in shared physical and social environments rather than treated as a text interface with speech added at its edges~\cite{porcheron2018voice}. A household assistant should therefore support hands-free activation, prompt spoken feedback, and controlled multi-turn conversation. It must also accommodate routines, device names, and usage practices that differ across homes, while providing extension mechanisms that remain manageable for personal deployment. Unlike an individual work assistant, a shared home interface may further need to interpret requests with context appropriate to different family members.

\texttt{Homebot} is a locally deployable AI agent for conversational household assistance and automation. 
Voice, Telegram, and Feishu channels normalize requests through a shared Message Bus and Agent Runtime, which can combine language-model reasoning with registered tools and task-specific skills. This common path does not merge all history: messaging context is scoped to a channel and chat identifier, whereas every wake-word activation starts a bounded voice session. The voice channel uses local wake-word detection, streaming speech recognition, and incremental text-to-speech playback. A structured \texttt{dialogue\_state} specifies whether playback should end the interaction, collect a follow-up answer, or retain a continuous multi-turn exchange. Optional speaker verification supplies a member or guest signal to voice-request context; profile-conditioned long-term memory and member-specific skills remain planned extensions.

This report presents \texttt{Homebot}'s compact multi-channel architecture, explicit voice turn-control design, and extension contracts for channels, tools, and skills. Together, these design choices provide a practical foundation for customizable, multi-member home assistance without proposing new speech or language models.

\section{\texttt{Homebot} Design}

\begin{figure*}[t]
  \centering
  \includegraphics[width=0.9\textwidth]{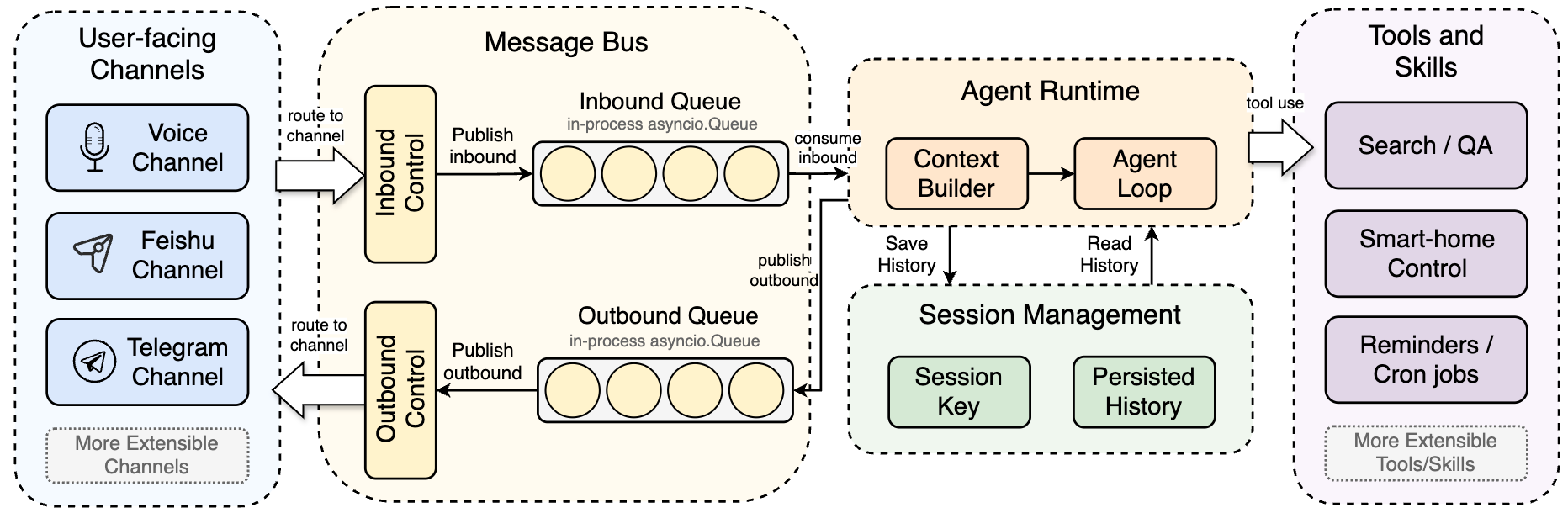}
  \caption{System overview of \texttt{Homebot}. User-facing channels route normalized messages through the message bus to the agent runtime, which constructs context, invokes tools and skills, and retrieves or persists channel-scoped session history.}
  \Description{A left-to-right architecture diagram showing voice, Feishu, and Telegram channels connected through inbound and outbound message-bus controls and queues to an agent runtime. The runtime contains a context builder and agent loop, connects to session management for persisted history, and invokes search, smart-home, and reminder capabilities from a tools-and-skills layer.}
  \label{fig:architecture}
\end{figure*}

\subsection{Overview}

Figure~\ref{fig:architecture} presents \texttt{Homebot} as a layered request-processing architecture with five responsibilities: \emph{Channels}, a \emph{Message Bus}, an \emph{Agent Runtime}, \emph{Session Management}, and \emph{Tools and Skills}. Channels form the system boundary: voice, Telegram, and Feishu translate platform-specific user events into a common message representation and render the resulting response in the appropriate modality. The Message Bus transports these normalized inbound messages to the runtime and routes outbound responses back to their originating channel. This separation lets all supported channels reuse one request path without exposing platform protocol details to the agent.

At the center of the architecture, the Agent Runtime constructs request context, invokes the language model, and interleaves model reasoning with registered tool calls. It draws executable capabilities from the Tools and Skills layer, which separates directly callable actions from task-specific instructions and constraints. Section~\ref{sec:agent-runtime} describes this execution path and its tool-use loop, while Section~\ref{sec:extensibility} details the corresponding extension contracts.

Session Management is deliberately outside this execution path: it determines whether a request may reuse prior conversational history, whereas the runtime processes the current request. Messaging sessions are scoped to a channel and chat identifier, so the common message path does not create cross-channel history by default (Section~\ref{sec:session-management}). Voice follows the same routing architecture after speech recognition, but a wake-word activation establishes a bounded interaction session; its wake-word, recognition, playback, and dialogue-state behavior are described in Section~\ref{sec:voice-interaction}.

\subsection{Agent Runtime}
\label{sec:agent-runtime}

Each channel converts an incoming event into a common message that records its channel, chat scope, content, and available metadata. This boundary prevents the Agent Runtime from depending on platform protocols and lets a new channel reuse the same task-processing path after implementing input normalization and reply delivery. The Message Bus keeps inbound receipt and outbound delivery separate, allowing channel listening, agent execution, and reply presentation to progress independently without introducing distributed-system machinery.

Before model execution, the \emph{Context Builder} assembles system identity and platform policy, the current request, runtime metadata, session history, and skill-related context. \texttt{Homebot} applies \emph{progressive disclosure} to skills: skills marked \texttt{always} contribute their complete instructions, whereas other usable skills are represented initially by a compact manifest containing their name, description, and location. A skill is task knowledge and operating guidance rather than an executable tool. When a task makes a skill relevant, the model uses the registered \texttt{read\_file} tool to retrieve its corresponding \texttt{SKILL.md} on demand; the retrieved instructions are returned as a tool result for the next reasoning step. This two-stage representation keeps broadly applicable guidance immediately available while deferring specialized instructions until needed, reducing per-turn token consumption and preserving effective context capacity for the active interaction.

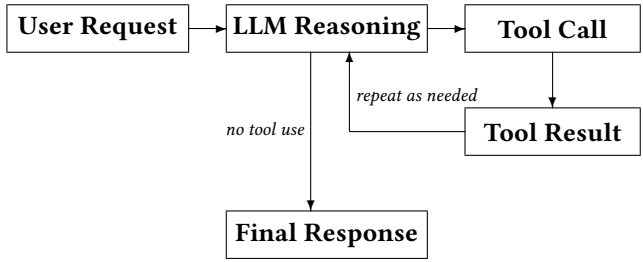
\begin{figure}[t]
  \centering
  \setlength{\unitlength}{0.001\columnwidth}
  \begin{picture}(1000,430)
    \put(0,320){\framebox(280,72){\textbf{User Request}}}
    \put(340,320){\framebox(310,72){\textbf{LLM Reasoning}}}
    \put(710,320){\framebox(270,72){\textbf{Tool Call}}}
    \put(710,160){\framebox(270,72){\textbf{Tool Result}}}
    \put(340,0){\framebox(310,72){\textbf{Final Response}}}

    \put(280,356){\vector(1,0){60}}
    \put(650,356){\vector(1,0){60}}
    
    \put(845,320){\vector(0,-1){88}}

    \put(710,196){\line(-1,0){180}} 
    \put(530,196){\line(0,1){84}}   
    \put(530,280){\vector(0,1){40}} 

    \put(470,320){\vector(0,-1){248}}

    \put(540,250){\makebox(0,0)[l]{\scriptsize\textit{repeat as needed}}}
    \put(460,200){\makebox(0,0)[r]{\scriptsize\textit{no tool use}}}
  \end{picture}
  \caption{The runtime returns tool results to model reasoning and produces a final response only when no further tool use is requested.}
  \Description{A LaTeX control-flow diagram with five nodes.}
  \label{fig:tool-loop}
\end{figure}

The \emph{Agent Runner} then performs an iterative request--action loop, illustrated in Figure~\ref{fig:tool-loop}. The model either produces a final response directly or issues a tool call; each tool result is returned to the model as additional context before the next reasoning step. On every model request, it supplies definitions for all currently registered tools; it does not remove tools based on a predicted intent. If the model requests an action, \texttt{Homebot} verifies the tool name, converts and validates parameters against the tool schema, executes the action, and appends a normalized result to the message history. The runner repeats this process until the model returns no further tool call, at which point its text is delivered as the user-facing response. Thus, a factual question may finish immediately, while a request such as scheduling a reminder can require one or more tool interactions. A configurable bound prevents unbounded action attempts; the default maximum is 200 iterations.

\subsection{Session Management}
\label{sec:session-management}

Session Management defines the lifecycle of conversational history rather than the execution of an individual request. Chat sessions use a key of the form \texttt{channel:chat\_id}. Consequently, different channels and chat scopes are isolated by default. For example, a Feishu direct message uses the sender as its chat identity, while group messages use the group identity. The Session Manager persists metadata and messages as JSONL files in the workspace \texttt{sessions/} directory; subsequent requests in the same scope can therefore resolve follow-up references from prior turns. The exact \texttt{/new} command cancels active work for the current session, clears and saves its messages, and invalidates the in-memory cache. It resets history without changing the session key or creating a cross-channel identity.

Voice uses a shorter-lived boundary. Each wake-word activation creates a fresh key of the form \texttt{voice:<UUID>}. Turns that remain within the active interaction can reuse this context, but the session is cleaned up after normal completion, timeout, an exit command, or cancellation. This model avoids silently carrying a previous spoken exchange into a later, independent activation while still supporting bounded multi-turn speech interaction.

\subsection{Voice Interaction Pipeline}
\label{sec:voice-interaction}

Voice is a complete interaction pipeline, rather than speech input and output around a text agent. While idle, the channel continuously applies local keyword spotting to microphone audio. A wake-word match plays a brief acknowledgement, creates the interaction-scoped voice session described above, and starts streaming speech recognition. The final transcript becomes a normalized inbound message, so voice requests use the same Agent Runtime, tools, and skills as text requests. This local front end limits full speech processing and model execution to explicit interactions.

The reply path is likewise incremental. The runtime requests a structured JSON voice response with a textual \texttt{reply} and a \texttt{dialogue\_state}. As reply text arrives, the channel splits it at sentence boundaries and feeds complete sentences to streaming text-to-speech. It can therefore begin playback before the entire response has been generated. The final structured response supplies the turn-control state after playback completes. \texttt{Homebot} integrates these components, but does not train wake-word, speech-recognition, speaker-recognition, or language models.

\subsubsection{Dialogue State and Turn Control}
\label{sec:dialogue-state}

\begin{figure}[t]
  \centering
  \setlength{\unitlength}{0.001\columnwidth}
  \begin{picture}(1000,600)
    \put(350,535){\framebox(300,60){\textbf{STOPPED}}}
    \put(350,400){\framebox(300,60){\textbf{LISTENING}}}
    \put(350,265){\framebox(300,60){\textbf{RECOGNIZING}}}
    \put(350,130){\framebox(300,60){\textbf{THINKING}}}
    \put(350,0){\framebox(300,60){\textbf{PLAYING}}}

    \put(500,535){\vector(0,-1){75}}
    \put(500,400){\vector(0,-1){75}}
    \put(500,265){\vector(0,-1){75}}
    \put(500,130){\vector(0,-1){70}}

    \put(515,495){\makebox(0,0)[l]{\scriptsize\textit{start}}}
    \put(515,360){\makebox(0,0)[l]{\scriptsize\textit{wake word}}}
    \put(515,225){\makebox(0,0)[l]{\scriptsize\textit{final spoken request}}}
    \put(515,90){\makebox(0,0)[l]{\scriptsize\textit{first speakable reply}}}

    \put(350,295){\line(-1,0){120}}
    \put(230,295){\line(0,1){125}}
    \put(230,420){\vector(1,0){120}}
    \put(225,357){\makebox(0,0)[r]{\scriptsize\shortstack[r]{\textit{silence} \\ \textit{timeout}}}}

    \put(350,30){\line(-1,0){270}}
    \put(80,30){\line(0,1){410}}
    \put(80,440){\vector(1,0){270}}
    \put(70,235){\makebox(0,0)[r]{\scriptsize\textit{end}}}

    \put(650,30){\line(1,0){230}}
    \put(880,30){\line(0,1){265}}
    \put(880,295){\vector(-1,0){230}}
    \put(890,162){\makebox(0,0)[l]{\scriptsize\shortstack[l]{\textit{follow\_up} \\ \textit{/ continuous}}}}
  \end{picture}
  \caption{Voice interaction state machine. The structured dialogue state determines the transition after playback.}
  \Description{A state machine with five vertically arranged nodes.}
  \label{fig:voice-state}
\end{figure}
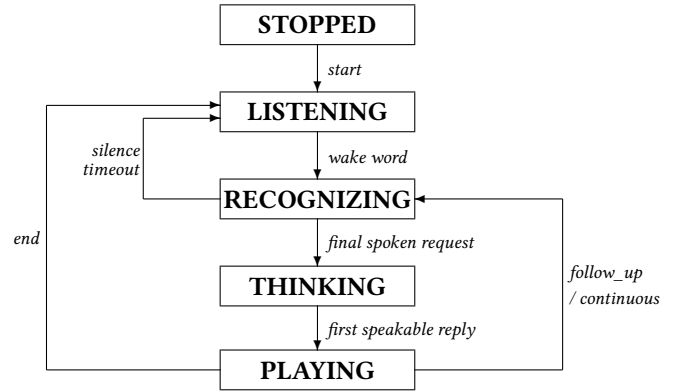

For reliable hands-free conversation, speech interaction requires an explicit decision about what happens after an answer. \texttt{Homebot} separates this decision from response text through a \texttt{dialogue\_\allowbreak state} field with three values:
\begingroup
\setlength{\leftmargini}{1.2em}
\begin{itemize}
  \item \texttt{end}: ends the current activation and returns the channel to wake-word listening;
  \item \texttt{follow\_up}: restarts recognition with the ordinary silence timeout when the system expects another answer;
  \item \texttt{continuous}: enters a persistent multi-turn mode, for example when a user explicitly asks to discuss a topic. Once entered, this mode remains active until an exit command, such as ``goodbye,'' or the longer silence timeout ends the interaction.
\end{itemize}
\endgroup
This protocol assigns semantic judgment to the model while leaving state transitions to the channel. For example, a request to set an alarm without a time produces a clarifying question and the \texttt{follow\_up} state, so recognition remains available for the user's answer.

Figure~\ref{fig:voice-state} summarizes the five user-visible channel states. The channel starts in \texttt{STOPPED} and enters \texttt{LISTENING} when started. A wake word moves it to \texttt{RECOGNIZING}; a final spoken request then moves it to \texttt{THINKING}. The first speakable reply starts \texttt{PLAYING}. After playback, \texttt{follow\_up} resumes recognition, while \texttt{end} returns to listening unless continuous mode is already active. An exit command, cancellation, or a silence timeout also clears the current interaction and returns to listening. The separation between incremental playback and final state parsing permits prompt speech without making the next-state decision premature.

\subsubsection{Speaker Recognition and Family Personalization}
\label{sec:voice-personalization}

Most personal home assistants implicitly assume one primary user, even though a household commonly includes several people with different routines, preferences, and references to rooms or devices. Optional speaker verification gives \texttt{Homebot} a lightweight member signal for a shared voice interface. It derives a speaker embedding from a bounded segment of the user's speech and compares it by cosine similarity with enrolled samples. The highest-scoring member is attached to the request when its score meets the configured threshold; otherwise, the request is labeled \texttt{guest}. Multiple enrollment samples per member allow the comparison to use the closest available sample.

The resulting member identifier enters voice-request metadata and can condition the Agent Runtime's current context. This enables \texttt{Homebot} to distinguish shared household context, such as device aliases and schedules, from member-specific context. The current implementation passes the member signal to the runtime. A planned profile-conditioned memory layer would retrieve and write long-term preferences, routines, and personal device aliases for the recognized member, while retaining shared home memory for requests that concern the household. The same boundary can later support member-specific skills and other personalized capabilities.

\subsection{Extensibility}
\label{sec:extensibility}

\texttt{Homebot} exposes three extension boundaries. \emph{Channels} implement a common lifecycle and sending contract, normalize platform events into inbound messages, and can optionally stream output. Built-in implementations are scanned from the package and external implementations can be discovered through Python entry points, but only channels configured as enabled are started. \emph{Tools} are explicitly registered executable capabilities. Representative implementations read workspace files, send messages through the bus, search current public information, and automate a persistent browser when those optional capabilities are configured. A custom tool must declare its name, description, JSON Schema parameters, and asynchronous execution method, then be explicitly registered in the \texttt{ToolRegistry}; placing a file in a directory does not make it callable. Finally, \emph{Skills} package task knowledge and operating constraints in \texttt{SKILL.md} files. Workspace skills take precedence over built-in skills with the same name, and dependency requirements determine whether a skill is available. Together, these boundaries keep channel adaptation, executable actions, and task guidance extensible without changing the shared runtime path.

\section{Conclusion}

\texttt{Homebot} shows how a personal home assistant can unify voice and messaging channels behind a shared agent runtime while preserving modality-specific session boundaries. Its voice design combines wake-word activation, streaming speech, and explicit turn control to support bounded multi-turn interaction. The resulting channel, tool, and skill contracts provide an extensible foundation for locally deployable home assistance.

\bibliographystyle{ACM-Reference-Format}
\bibliography{references}

\end{document}